\documentclass[conference]{IEEEtran}
\IEEEoverridecommandlockouts
\usepackage{cite}
\usepackage{amsmath,amssymb,amsfonts}
\usepackage{algorithmic}
\usepackage{graphicx}
\usepackage{textcomp}
\usepackage{xcolor}
\usepackage{dirtytalk}
\usepackage{booktabs} 
\usepackage{multirow,tabularx}
\ifCLASSOPTIONcompsoc
 \usepackage[caption=false,font=footnotesize,labelfont=sf,textfont=sf]{subfig}
\else
 \usepackage[caption=false,font=footnotesize]{subfig}
\fi

\usepackage{tikz}
\newcommand*\circled[1]{\tikz[baseline=(char.base)]{
            \node[shape=circle,draw,inner sep=1pt] (char) {#1};}}

\newcommand\blfootnote[1]{%
  \begingroup
  \renewcommand\thefootnote{}\footnote{#1}%
  \addtocounter{footnote}{-1}%
  \endgroup
}

\def\BibTeX{{\rm B\kern-.05em{\sc i\kern-.025em b}\kern-.08em
    T\kern-.1667em\lower.7ex\hbox{E}\kern-.125emX}}
\begin{document}

\title{Towards Dependable Autonomous Systems\\Based on Bayesian Deep Learning Components}

\author{
\IEEEauthorblockN{
Fabio Arnez\IEEEauthorrefmark{1},
Huascar Espinoza\IEEEauthorrefmark{2},
Ansgar Radermacher\IEEEauthorrefmark{1} and 
Fran\c{c}ois Terrier\IEEEauthorrefmark{1}}

\IEEEauthorblockA{\IEEEauthorrefmark{1}Universit\'e Paris-Saclay, CEA, List, F-91120, Palaiseau, France\\
\{name.lastname\}@cea.fr}
\IEEEauthorblockA{\IEEEauthorrefmark{2}KDT JU, TO 56 05/16, B-1049 Brussels, Belgium\\
huascar.espinoza@kdt-ju.europa.eu}
}

\maketitle

\blfootnote{\textbf{Preprint version.} \textbf{Accepted} and presented at the $18^{th}$ European Dependable Computing Conference (EDCC), Zaragoza, Spain, 2022. Digital Object Identifier (DOI) is available in the preprint description.}

\begin{abstract}
As autonomous systems increasingly rely on Deep Neural Networks (DNN) to implement the navigation pipeline functions, uncertainty estimation methods have become paramount for estimating confidence in DNN predictions. Bayesian Deep Learning (BDL) offers a principled approach to model uncertainties in DNNs. However, in DNN-based systems, not all the components use uncertainty estimation methods and typically ignore the uncertainty propagation between them.
This paper provides a method that considers the uncertainty and the interaction between BDL components to capture the overall system uncertainty. We study the effect of uncertainty propagation in a BDL-based system for autonomous aerial navigation. Experiments show that our approach allows us to capture useful uncertainty estimates while slightly improving the system's performance in its final task. In addition, we discuss the benefits, challenges, and implications of adopting BDL to build dependable autonomous systems.
\end{abstract}


\section{Introduction}
Navigation in complex environments still represents a big challenge for autonomous systems (AS). Particular instances of this problem are autonomous driving and autonomous aerial navigation in the context of self-driving cars and Unmanned Aerial Vehicles (UAVs), respectively. In both cases, the navigation task is addressed by first acquiring rich and complex raw sensory information (e.g., from camera, radar, LiDAR, etc.), which is then processed to drive the autonomous agent towards its goal.
Usually, this process is done in sequence, where tasks and specific software components are linked together in the so-called \textit{perception-planning-control} software pipeline \cite{grigorescu2020survey, mcallister2017concrete}.


Over the last decade, Deep Neural Networks (DNNs) have become a popular choice to implement navigation pipeline components thanks to their effectiveness in processing complex sensory inputs, and their powerful representation learning that surpasses the performance of traditional methods. Currently, three main paradigms exist to develop and train navigation components based on DNNs: Modular (isolated), End-to-End (E2E) learning, and mixed or hybrid approaches~\cite{mcallister2017concrete}.




\begin{figure}[!t]
	\centering
	\includegraphics[width=0.95\linewidth]{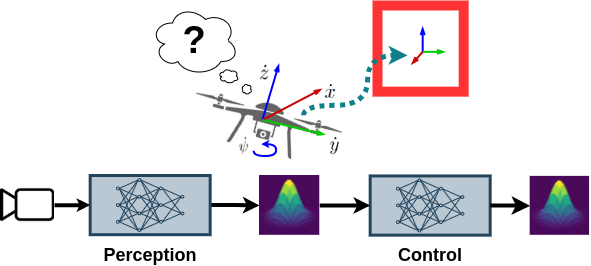}
	\caption{UAV BDL-based Aerial Navigation Pipeline: The downstream control component gets predictions of the previous perception component as input and must take their uncertainty into account.}
	\label{fig:simple-uncertainty-nav}
\end{figure}

Despite the remarkable progress in representation learning, DNNs should also represent the confidence in their predictions to deploy them in safety-critical systems. McAllister et al. \cite{mcallister2017concrete} proposed using Bayesian Deep Learning (BDL) to implement the components from navigation pipelines or stacks. Bayesian methods offer a principled framework to model and capture system uncertainty. However, if the Bayesian approach is followed, all the components in the system pipeline should use BDL to enable uncertainty propagation in the pipeline. Hence, BDL components should admit uncertainty information as an input to account for the uncertainty from the outputs of preceding BDL components (See Fig.~\ref{fig:simple-uncertainty-nav}).

In recent years, a large body of literature has employed uncertainty estimation methods in robotic tasks thanks to its potential to improve the safety of automated functions \cite{michelmore2018evaluating}, and the capacity to increase the task performance \cite{nozarian2020uncertainty, ohn2020learning}. However, uncertainty is captured partially in navigation pipelines that utilize DNNs. BDL methods are used mainly in perception tasks, and downstream components (e.g., planning and control) usually ignore the uncertainty from the preceding components or do not capture uncertainty in their predictions.

Although some works propagate downstream perceptual uncertainty from intermediate representations \cite{arnez2021improving,ivanovic2021heterogeneous,casas2020implicit}, the overall system output does not take into account all the uncertainty sources from DNN components in the pipeline. Moreover, proposed frameworks for dynamic dependability management that use uncertainty information focus only on DNN-based perception tasks \cite{henne2019managing, asaadi2020quantifying}, ignoring uncertainty propagation through the system pipeline, the interactions between uncertainty-aware components, and the potential impact on system performance and safety.


Quantifying uncertainty in a BDL-based system (i.e., a pipeline of BDL components) still remains a challenging task. Uncertainties from BDL components must be assembled in a principled way to provide a reliable measure of overall system uncertainty, based on which safe decisions can be made \cite{mcallister2017concrete,lavin2021technology}. In this paper, we propose to capture the uncertainty along a pipeline of BDL components and study the impact of uncertainty propagation on the aerial navigation task in a UAV. In addition, we propose an uncertainty-centric dynamic dependability management framework to cope with the challenges that arise from propagating uncertainty through BDL-based systems.

\section{Related Work}

\subsection{Neural Network Uncertainty Estimation}

Bayesian neural networks (BNN) have been widely used to represent the confidence in the predictions. A proper confidence representation in DNN predictions can be achieved by modeling two sources of uncertainty: \textit{aleatoric} (data) and \textit{epistemic} (model) uncertainty. For \textit{epistemic} uncertainty, Bayesian inference is used to estimate the posterior predictive distribution. In practice, approximate Bayesian inference methods are often used \cite{gal2016dropout,gal2017concrete, lakshminarayanan2017simple,gustafsson2019evaluating} since the posterior on the model parameters $p(\theta \mid \mathcal{D})$ is intractable in DNNs.

To model data uncertainty, \cite{kendall2017uncertainties, lakshminarayanan2017simple} propose to incorporate additional outputs to represent the parameters (mean and variance) of a Gaussian distribution. Loquercio et al.\cite{loquercio2020general} forward propagate sensor noise through the DNN. This approach does not require retraining, however, it assumes a fixed uncertainty value for the sensor noise at the input. Another family of methods aim to capture complex stochastic patterns such as multimodality or heteroscedasticity (\textit{aleatoric} uncertainty) using latent variables (LV) as input.  When BNNs are used with LV (BNN+LV), both types of uncertainty can be captured \cite{depeweg2017learning,  depeweg2018decomposition}. In this approach, a BNN receives an input combined with a random disturbance coming from an LV (i.e., features are partially stochastic). In contrast, this paper considers that a BNN can receive a complete stochastic features at the input.

\subsection{Uncertainty in DNN-Based Navigation}

In an autonomous driving context, perception uncertainty is captured from implicit \cite{casas2020implicit} and explicit representations \cite{ivanovic2021heterogeneous} and used downstream for scene motion forecasting and trajectory planning respectively. In reinforcement learning, input uncertainty has been employed for model-based \cite{lutjens2019safe} and model-free control policies \cite{fan2020learning}. In the former case, a collision predictor uncertainty is passed to a model predictive controller. In the latter, perception uncertainty is mapped to the control policy uncertainty using heuristics. In the context of aerial navigation, a few works have considered uncertainty. \cite{loquercio2020general} uses a fixed uncertainty value for sensors as an input to a control policy. \cite{arnez2021improving} extends the work from \cite{bonatti2019learning} to use the uncertainty from perception noisy representations downstream in a BNN control policy. Although these approaches use perception uncertainty in downstream components, not all the DNN components in the pipeline employ uncertainty estimation methods.

\subsection{Uncertainty-based Dependability Frameworks}
\label{subsec:uncertainty-dependability-fw}
For the deployment of dependable autonomous systems that use machine learning
(ML) components, Trapp et al. \cite{trapp2018towards} and Henne et al.\cite{henne2019managing} conceptualized the use and runtime monitoring of perception uncertainty to ensure safe behavior on AS. To model system behavior, probabilistic graphical models (PGMs) and, in particular, Bayesian Networks (BNs) have been used in dependability research for safety and reliability analyses and risk assessment applications \cite{kabir2019applications}. BNs allow incorporating expert domain knowledge, model complex relationships between components, and enable decision-making under uncertainty. In the context of autonomous aviation systems, \cite{asaadi2020quantifying} proposes a method for quantifying system assurance using perception component uncertainty and dynamic BNs. For autonomous vehicles, \cite{reich2021towards} offers a framework for dynamic risk assessment, using BNs to predict the behavior intents of other traffic participants. Unlike these works, this paper considers uncertainty from Bayesian deep learning components beyond perception.

\section{System Task Formulation}

\begin{figure}[!t]
	\centering
	\includegraphics[width=1.0\linewidth]{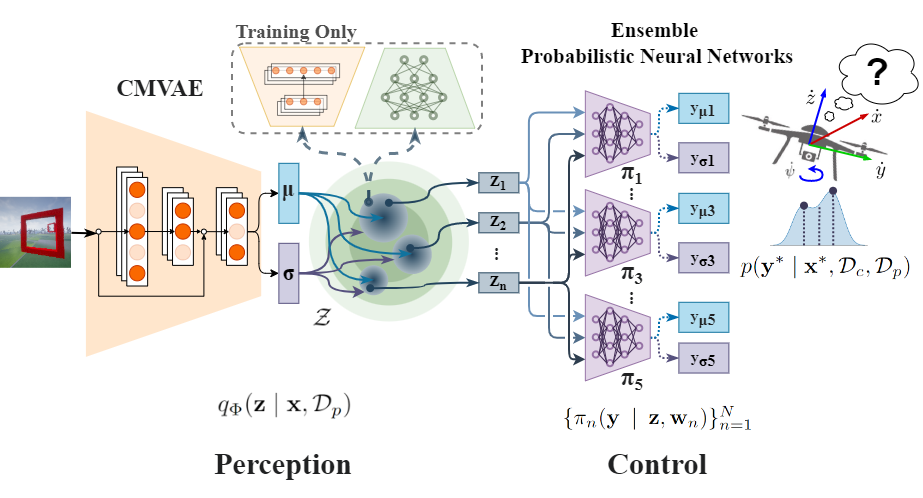}
	\caption{System architecture for aerial navigation}
	\label{fig:SystemArch}
\end{figure}

In this paper, we address the problem of autonomous aerial navigation. The goal of the autonomous agent (i.e., UAV) is to navigate through a set of gates with unknown locations disposed in a circular track. Following prior work from \cite{bonatti2019learning,arnez2021improving}, the navigation architecture consists of two DNN-based components: one for perception and the other for control (see Fig.~\ref{fig:SystemArch}). Both DNNs are trained following the hybrid paradigm. To achieve the agent goal, the navigation task is formulated as a sequential-decision making problem, where a sequence of control actions are produced given environment observations. In this regard, the simulation environment provides at each time step an observation comprised of an RGB image $\mathbf{x}$ acquired from a front-facing camera on the UAV. The perception component defines an encoder function  $q_{\phi}:\mathcal{X} \rightarrow \mathcal{Z}$ that maps the input image $\mathbf{x}$ to a rich low dimensional representation  $\mathbf{z} \in \mathbb{R}^{10}$. Next, a control policy $\pi_{w}: \mathcal{Z} \rightarrow \mathcal{Y}$ maps the compact representation $\mathbf{z}$ to control commands $\mathbf{y} = [\dot{x}, \dot{y}, \dot{z}, \dot{\psi}] \in \mathbb{R}^{4}$, corresponding to linear and angular (yaw) velocities in the UAV body frame.

In the perception component, a cross-modal variational autoencoder (CMVAE) \cite{spurr2018cross,bonatti2019learning} is used to learn a rich and robust compact representation. A CMVAE is a variant of the traditional variational autoencoder (VAE) \cite{kingma2013auto} that learns a single latent representation for multiple data modalities. In this case, the perception dataset $\mathcal{D}_p$ has two data modalities: the RGB images and the pose of the gate relative to the UAV body-frame. During training, the CMVAE encoder $q_{\phi}$ maps an input image $\mathbf{x}$ to a noisy representation with mean $\mu_{\phi}(\mathbf{x})$ and variance $\sigma^{2}_{\phi}(\mathbf{x})$ in the latent space, from where latent vectors $\mathbf{z}$ are sampled, $\mathbf{z} \sim \mathcal{N}(\mu_{\phi},\sigma^{2}_{\phi})$. Next, a latent vector $\mathbf{z}$ is used to reconstruct the input image and estimate the gate pose (i.e., recover the two data modalities) using two DNNs, a decoder and a feed-forward network. The CMVAE encoder $q_{\phi}$ is based on the Dronet architecture \cite{loquercio2018dronet}, and additional constraints on the latent space are imposed through the loss function to promote the learning of robust disentangled representations.

Once the perception component is trained, the downstream control task (control policy $\pi$) uses a feed-forward network to operate on the latent vectors $\mathbf{z}$ at the output of the CMVAE encoder $q_{\phi}$ to predict UAV velocities. To this end, the control policy network is added at the output of the perception encoder $q_{\phi}$, forming the navigation pipeline DNN. The control component $\pi$ uses a control imitation learning dataset ($\mathcal{D}_c$). During training, we freeze the perception encoder $q_{\phi}$ to update only the control policy network. For more information about the general architecture for aerial navigation, datasets, and training procedures, we refer the reader to \cite{bonatti2019learning,arnez2021improving}.

\section{Methodology}

\subsection{Uncertainty from Perception Representations}

Although the CMVAE encoder $q_{\phi}$ employs Bayesian inference to obtain latent vectors $\mathbf{z}$, CMVAE does not capture epistemic uncertainty since the encoder lacks a distribution over parameters $\phi$. To capture uncertainty in the perception encoder we follow prior work from \cite{daxberger2019bayesian,jesson2020identifying} that attempts to capture epistemic uncertainty in VAEs. We adapt the CMVAE to capture the posterior $q_\Phi(\mathbf{z} \mid \mathbf{x}, \mathcal{D}_p)$ as shown in (\ref{eq:postEncoder}).

\begin{equation}
    q_{\Phi}(\mathbf{z} \mid \mathbf{x}, \mathcal{D}_{p}) = \int{q(\mathbf{z} \mid \mathbf{x}, \phi) p(\phi \mid \mathcal{D}_{p})d\phi}
	\label{eq:postEncoder}
\end{equation}

To approximate (\ref{eq:postEncoder}), we take a set ${\Phi = \{\phi_{m}\}^{M}_{m}}$ of encoder parameters samples $\phi_{m} \sim p(\phi \mid \mathcal{D}_{p})$, to obtain a set of latent samples $\{\mathbf{z}_{m}\}^{M}_{m=1} \sim q_{\Phi}(\mathbf{z} \mid \mathbf{x}, \mathcal{D}_{p})$ at the output of the encoder. In practice, we modify CMVAE by adding a dropout layer in the encoder. Then, we use Monte Carlo Dropout (MCD) \cite{gal2016dropout} to approximate the posterior on the encoder weights $p(\phi \mid \mathcal{D}_{p})$. Finally, for a given input image $\mathbf{x}$ we perform $M$ stochastic forward passes (with dropout \say{turned on}) to compute a set of $M$ latent vector samples $\mathbf{z}$ at runtime.

\subsection{Input Uncertainty for Control}

In BDL, downstream uncertainty propagation assumes that a neural network component is able to handle or admit uncertainty at the input. In our navigation case, this implies that the DNN-based controller is able to handle the uncertainty coming from the perception encoder $q_{\Phi}$. To capture the navigation model uncertainty (overall system uncertainty at the output of the controller), we use the Bayesian approach to compute the posterior predictive distribution for target variable $\mathbf{y^*}$ associated with a new input image $\mathbf{x^*}$, as shown in (\ref{eq:postPredDist}).
\begin{multline}
    p(\mathbf{y^*} \mid \mathbf{x^*}, \mathcal{D}_{c}, \mathcal{D}_{p}) = \\
	        \iint{\pi(\mathbf{y} \mid \mathbf{z}, \mathbf{w}) p(\mathbf{w} \mid \mathcal{D}_{c})q_{\Phi}(\mathbf{z} \mid \mathbf{x^{*}}, \mathcal{D}_{p}) d\mathbf{w} d\mathbf{z}}
	\label{eq:postPredDist}
\end{multline}

The integrals from (\ref{eq:postPredDist}) are intractable, and we rely on approximations to obtain an estimation of the predictive distribution. The posterior $p(\mathbf{w} \mid \mathcal{D}_{c})$ is difficult to evaluate, thus we can approximate the inner integral using an ensemble of neural networks \cite{gustafsson2019evaluating}. In practice, we train an ensemble of $N$ probabilistic control policies $\{ \mathbf{\pi}_{n}(\mathbf{y} \mid \mathbf{z}, \mathbf{w}_{n})\}_{n=1}^{N}$, with weights $\{\mathbf{w}_{n}\}^{N}_{n=1} \sim p(\mathbf{w | \mathcal{D}})$. Each control policy $\pi_{n}$ in the ensemble predicts the mean $\mu$ and variance $\sigma^{2}$ for each velocity command, i.e.,
$\mathbf{y}_{\mu} = [\mu_{\dot{x}}, \mu_{\dot{y}}, \mu_{\dot{z}}, \mu_{\dot{\psi}}]$ and
$\mathbf{y}_{\sigma^{2}} = [\sigma^{2}_{\dot{x}}, \sigma^{2}_{\dot{y}}, \sigma^{2}_{\dot{z}}, \sigma^{2}_{\dot{\psi}}]$. For training the control policy we use imitation learning and the heteroscedastic loss function, as suggested by \cite{kendall2017uncertainties, lakshminarayanan2017simple}.

The outer integral is approximated by taking a set of samples from the perception component latent space. In \cite{arnez2021improving} latent samples are drawn using the encoder mean and variance $\mathbf{z} \sim \mathcal{N}(\mu_{\phi},\sigma^{2}_{\phi})$. For the sake of simplicity, we directly use the samples obtained in the perception component $\{\mathbf{z}_{m}\}^{M}_{m} \sim q_{\Phi}(\mathbf{z} \mid \mathbf{x}, \mathcal{D}_{p})$ to take into account the epistemic uncertainty from the previous stage. Finally, the predictions that we get from passing each latent vector $\mathbf{z}$ through each ensemble member are used to estimate the posterior predictive distribution in (\ref{eq:postPredDist}). From the control policy perspective, using multiple latent samples $\mathbf{z}$ can be seen as taking a better \say{picture} of the latent space (perception representation) to gather more information about the environment.

\section{Experiments \& Discussion}
For our experiments, we seek to study the impact of uncertainty propagation in the navigation pipeline. In particular, we seek to answer the following research questions: \textbf{RQ1.}~How does uncertainty from perception representations affect downstream component uncertainty estimation quality? \textbf{RQ2.}~Can uncertainty propagation improve system performance? \textbf{RQ3.}~Could uncertainty-aware components in the pipeline help detect challenging scenes that can threaten the system mission? To answer these questions we perform a quantitative and qualitative comparison between uncertainty-aware aerial navigation models. 


\subsection{Experimental setup}

\subsubsection{Navigation Model Baselines}
All the navigation architectures are based on \cite{bonatti2019learning} and are implemented using PyTorch. Table \ref{table:exp-nav-models} shows the uncertainty-aware navigation architectures used in our experiments, detailing the type of perception component, the number of latent variable samples (LVS), the type of control policy, and the number of control prediction samples (CPS) at the output of the system. For instance, $\mathcal{M}_0$ represents our Bayesian navigation pipeline. $\mathcal{M}_0$ perception component captures epistemic uncertainty using MCD with 32 forward passes for each input to get 32 latent variable predictions. For the sake of simplicity, perception predictions are directly used as latent variable samples in downstream control. The control component uses an ensemble of 5 probabilistic control policies obtaining 160 control prediction samples. $\mathcal{M}_1$ to $\mathcal{M}_4$ partially capture uncertainty in the pipeline since they use a deterministic perception component (CMVAE). For the control component, $\mathcal{M}_1$ and $\mathcal{M}_2$ take 32 and 1 latent variable samples (LVS) respectively, and use the samples later with an ensemble of 5 probabilistic control policies capturing epistemic and aleatoric uncertainty; $\mathcal{M}_3$ uses 32 LVS, and the control component is completely deterministic; $\mathcal{M}_4$ uses 1 LVS with a probabilistic control policy to capture aleatoric uncertainty. For UAV control, we use the expected value of the predicted velocities means at the output of the control component \cite{lakshminarayanan2017simple}, i.e.,~$\mathbf{\hat{y}}_{\mu} = \mathbb{E}([\mu_{\dot{x}}, \mu_{\dot{y}}, \mu_{\dot{z}}, \mu_{\dot{\psi}}])$.



\begin{table}[!t]
\caption{Uncertainty-aware Navigation models in the experiments}
\label{table:exp-nav-models}
\centering
\begin{tabular}{@{}cllll@{}}
\toprule
\textbf{Model} & \textbf{Perception $(q_{\phi})$} & \textbf{LVS} & \textbf{Control Policy $(\pi)$} & \textbf{CPS} \\ \midrule
$\mathcal{M}_0$ & MCD-CMVAE & 32 & Ensemble ($N=5$) Prob. & 160 \\
$\mathcal{M}_1$ & CMVAE     & 32 & Ensemble ($N=5$) Prob. & 160 \\
$\mathcal{M}_2$ & CMVAE     & 1  & Ensemble ($N=5$) Prob. & 5   \\
$\mathcal{M}_3$ & CMVAE     & 32 & Deterministic          & 32  \\
$\mathcal{M}_4$ & CMVAE     & 1  & Prob.          & 1   \\ \bottomrule
\end{tabular}
\end{table}

\subsubsection{Datasets}
We use two independent datasets for each component in the navigation pipeline. The perception CMVAE uses a dataset ($\mathcal{D}_p$) of 300k images where a gate is visible and gate-pose annotations area available. The control component uses a dataset ($\mathcal{D}_c$) of 17k images with UAV velocity annotations. $\mathcal{D}_c$ is collected by flying the UAV in a circular track with gates, using traditional methods for trajectory planning and control (see \cite{bonatti2019learning} for more details).
The perception dataset is divided into 80\% for training, and the remaining 20\% for validation and testing. The control dataset uses a split of 90\% for training and the remaining for validation and testing. In both cases the image size is 64x64 pixels. In addition, using the validation data from $\mathcal{D}_p$ and $\mathcal{D}_c$, we generate refined validation sub-datasets with images that have: exactly one visible gate (ideal situation), no visible gate in front, and multiple gates visible. The last two types of images represent situations that can pose a risk to the system task. Each sub-dataset contains 200 images.

\subsection{Experiments}
In the context of RQ2, we use the validation dataset from the control component to measure the regression Expected Calibration Error (ECE) \cite{kuleshov2018accurate} to compare the quality of uncertainty estimates from navigation models at the output of the system, (i.e., the control component output).

In order to answer RQ2, we evaluate our navigation architecture under controlled simulations using the AirSim simulation environment. The UAV mission resembles the scenario and the conditions observed in the training dataset. Therefore, we use a circular track with eight equally spaced gates positioned initially in a radius of 8m and constant height. To assess the system performance to perturbations in the environment, we generate new tracks adding random noise to each gate radius and height.

\begin{figure}[t!]
    \centering
        \subfloat[\centering Circular track view without noise (left) and with noise (right).]
            {
            \includegraphics[width=0.46\linewidth]{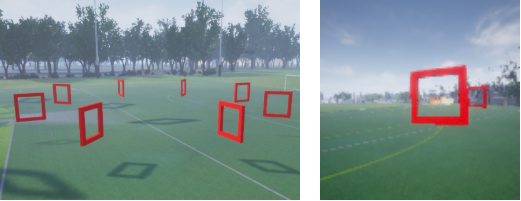}
            \quad
            \includegraphics[width=0.46\linewidth]{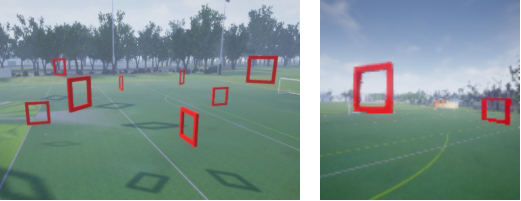}
            \label{subfig:tracks}
            }

            \qquad
            
        \subfloat[\centering UAV mission scenes]
        {
            \includegraphics[width=0.25\linewidth]{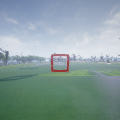}
            \quad
            \includegraphics[width=0.25\linewidth]{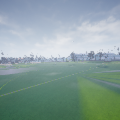}
            \quad
            \includegraphics[width=0.25\linewidth]{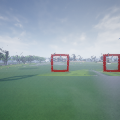}
            \label{subfig:scenes}
        }
            
    \caption{UAV Mission: Navigation tracks and scenes from birds-eye view, and view from UAV perspective}    
    
    \label{fig:TrackViews}
\end{figure}

In the context of the AirSim \cite{shah2018airsim} simulation environment, a track is entirely defined by a set of gates, their poses in three-dimensional space, and the expected navigation direction of the agent. For perception-based navigation, the complexity of a track resides in the \say{\textit{gate-visibility}} difficulty \cite{madaan2020airsim,song2021autonomous}, i.e., how well the camera Field-of-View (FoV) captures the gate. A natural way to increase track complexity is by adding a random displacement to the position of each gate. A track without random displacement in the gates has a circular fashion. Gate position randomness alters the shape of the track, affecting the gate visibility, i.e., gates are: not visible, partially visible, or multiple gates can be captured in the UAV FoV as presented in Fig.~\ref{fig:TrackViews}. The images from these scenarios are challenging given its potential impact on system performance.

To measure the system performance we take the average number of gates passed in all generated tracks. For track generation we use a random seed to produce circular tracks with two levels of noise in the gates offset, i.e., each random seed generates two (reproducible) noisy tracks. In total, we use 6 random seeds to produce 12 tracks, 6 tracks per noise level. The two noise levels are a combination of Gate Radius Noise (GRN) and Gate Height Noise (GHN). Finally, all navigation models are tested in the same generated tracks for a fair comparison, and each model has 3 trials per track.



To address RQ3, we perform a qualitative comparison of the component predicted densities using scenes (images) from challenging situations during the UAV mission. To this end, we first use the images from the generated sub-datasets. Next, we use the scenes from Fig. \ref{subfig:scenes} as an input to the Bayesian navigation model $\mathcal{M}_0$ to analyze the effect uncertainty propagation under specific situations.

\subsection{Results}
Table \ref{table:nav-models-exp-results} summarizes the ECE for all the navigation models using the validation dataset from the control component. $\mathcal{M}_4$ has the best uncertainty quality since the model learned to predict the noise from the data using the heteroscedastic loss function. On the contrary, $\mathcal{M}_2$ has the worst calibration results caused by the deterministic control choice and its inability to learn the data uncertainty. $\mathcal{M}_1$ and $\mathcal{M}_2$ have similar values since both receive the one noisy encoding from perception. However, $\mathcal{M}_1$ takes multiple samples from the noisy perception encoding which causes a reduction of the ECE value. Finally, $\mathcal{M}_0$ shows a higher ECE value compared to the previous models. This is caused by applying MCD in the perception CMVAE and the dispersion of the latent codes at the output of the perception encoder $q_{\Phi}$. The uncertainty quality of the downstream control is slightly affected because the control component did not see the same perception encoding dispersion (uncertainty) during training.

\begin{table}[!t]
\caption{Uncertainty-Aware Navigation Models:\\ECE \& Avg. number of gates passed}
\label{table:nav-models-exp-results}
\centering
\begin{tabular}{@{}cccc@{}}
\toprule
\multirow{2}{*}{\textbf{Model}} &
  \multirow{2}{*}{\textbf{ECE} ($\downarrow$)} &
  \multicolumn{2}{c}{\textbf{Performance with Track Gate Noise} ($\uparrow$)} \\ \cmidrule(l){3-4} 
 &
   &
  \multicolumn{1}{l}{\textbf{\begin{tabular}[c]{@{}l@{}}$GRN \sim \mathcal{U}[-1.0, 1.0)$\\ $GHN \sim \mathcal{U}[0, 2.0)$\end{tabular}}} &
  \multicolumn{1}{l}{\textbf{\begin{tabular}[c]{@{}l@{}}$GRN \sim \mathcal{U}[-1.5, 1.5)$\\ $GHN \sim \mathcal{U}[0, 3.0)$\end{tabular}}} \\ \midrule
$\mathcal{M}_0$ & 0.00700  & \textbf{19.77} & \textbf{9.22} \\
$\mathcal{M}_1$ & 0.00129 & 17.67 & 6.0 \\
$\mathcal{M}_2$ & 0.00136 & 17.33  & 4.0 \\
$\mathcal{M}_3$ & 0.05709  & 8.33 & 5.0 \\
$\mathcal{M}_4$ & \textbf{0.00050 } & 15.16  & 4.38  \\ \bottomrule
\end{tabular}
\end{table}


For RQ2, Table \ref{table:nav-models-exp-results} presents the navigation performance results for all the navigation models. In general, learning to predict uncertainty in the control component can boost the performance significantly. However, for $\mathcal{M}_3$, sampling from a noisy perception representation adds sufficient diversity to the downstream control predictions, resulting in better decisions than $\mathcal{M}_2$ in tracks with higher noise levels. In $\mathcal{M}_4$, the good performance suggests that the track noise observed at test time (lower noise level), resembles the data noise observed during the training of the single probabilistic model.

\begin{figure*}[t!]
\centering
    \subfloat[\centering Single gate prediction densities]
        {
        \includegraphics[width=0.32\linewidth]{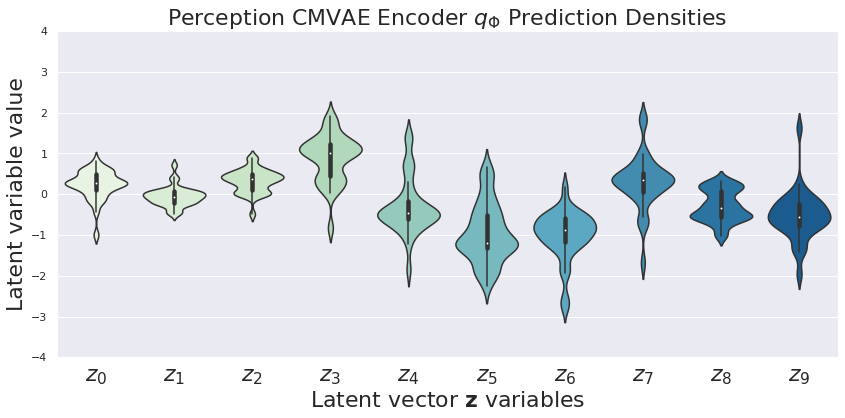}
        \hfil
        \includegraphics[width=0.32\linewidth]{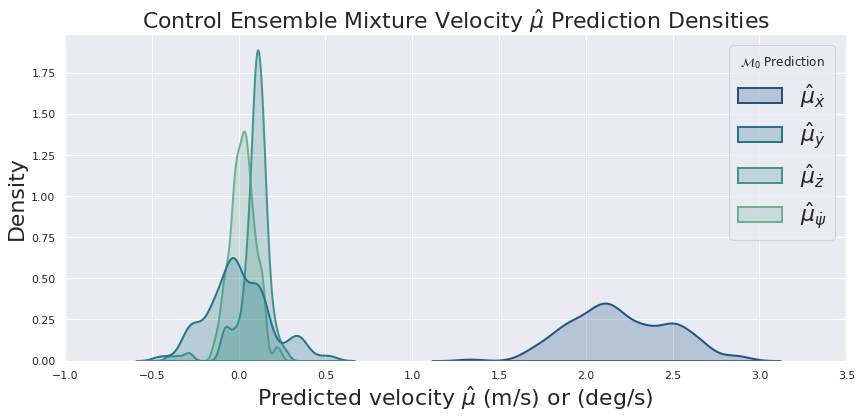}
        \hfil
        \includegraphics[width=0.32\linewidth]{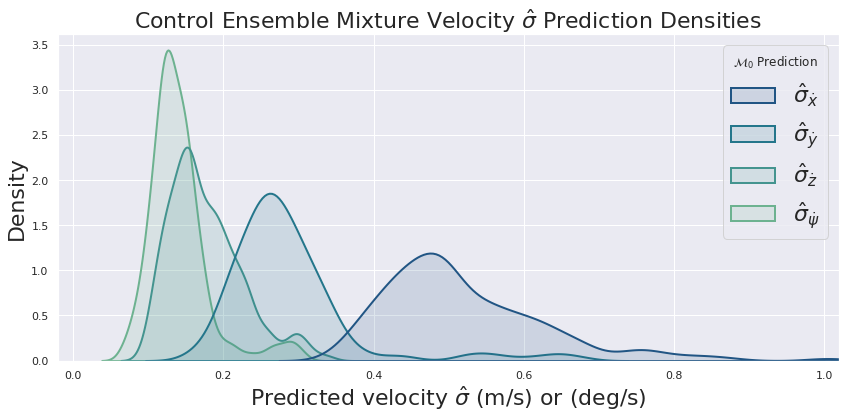}
        }
    
    \subfloat[\centering No visible gate prediction densities]
        {
        \includegraphics[width=0.32\linewidth]{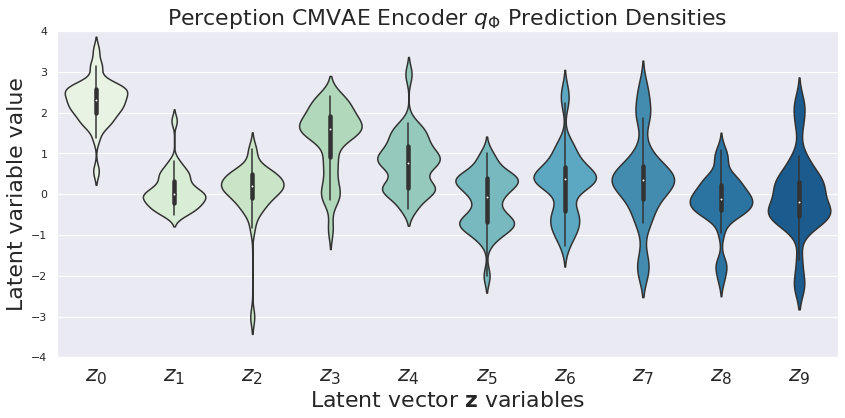}
        \hfil
        \includegraphics[width=0.32\linewidth]{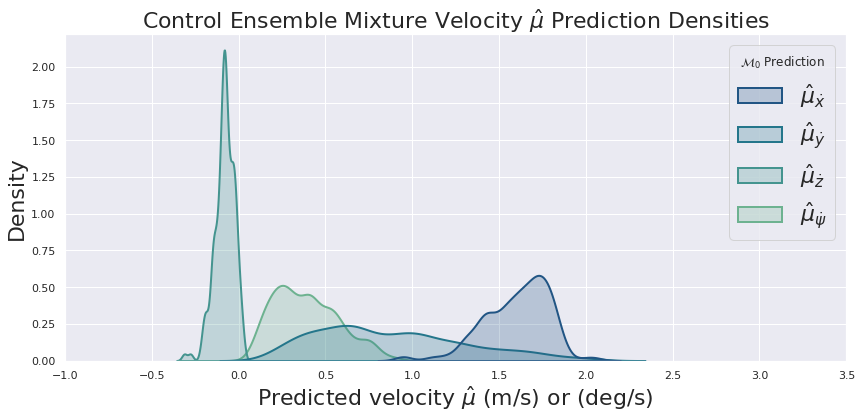}
        \hfil
        \includegraphics[width=0.32\linewidth]{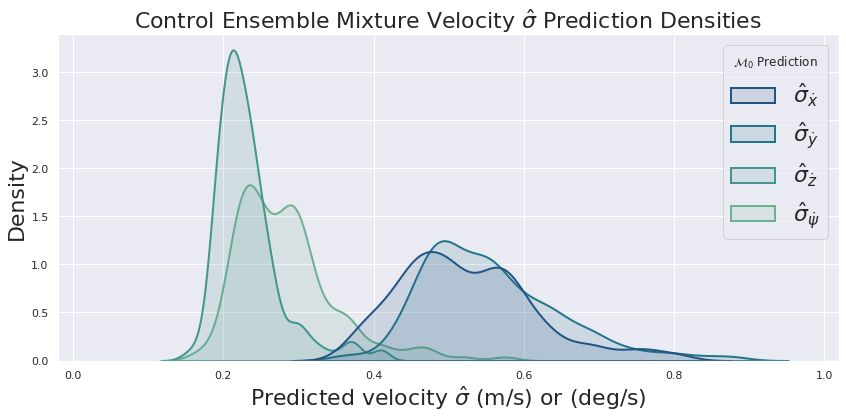}
        }
        
    \subfloat[\centering Double gate prediction densities]
        {
        \includegraphics[width=0.32\linewidth]{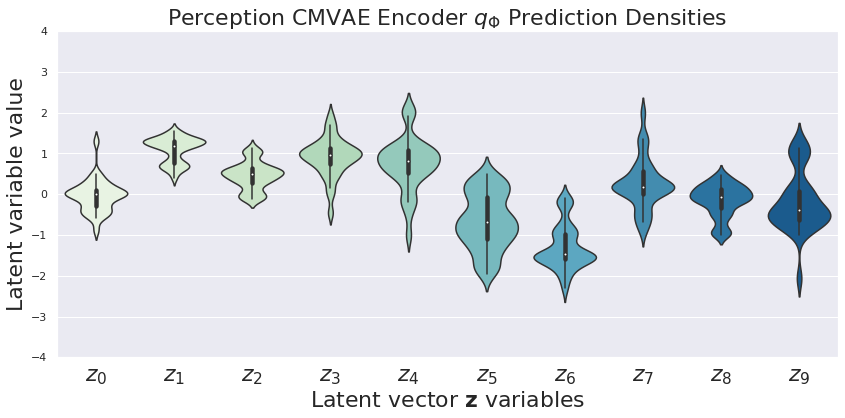}
        \hfil
        \includegraphics[width=0.32\linewidth]{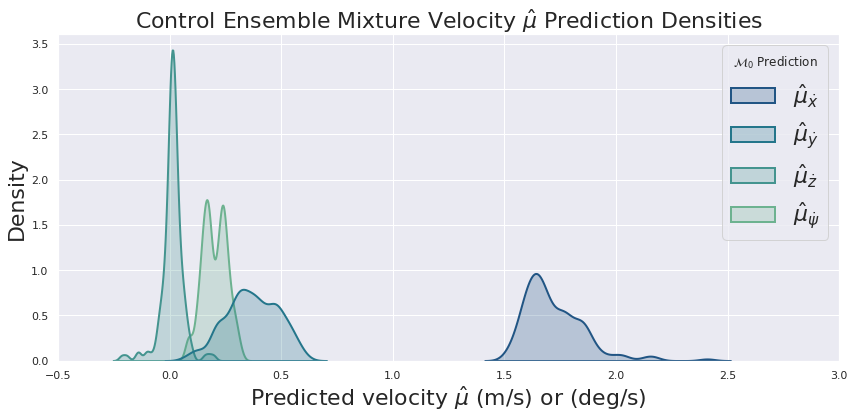}
        \hfil
        \includegraphics[width=0.32\linewidth]{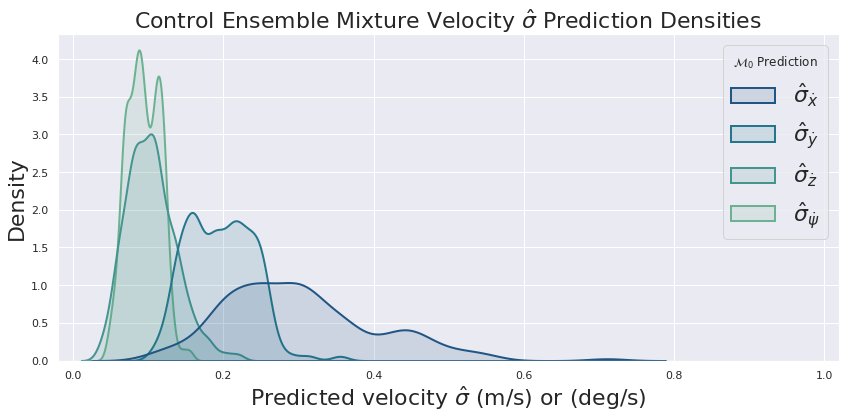}
        }
\caption{Bayesian navigation model $\mathcal{M}_0$: Perception $q_{\Phi}$ predictions $\mathbf{z}$ density (left); Predicted velocity $\hat{\mu}$ density (mid); Predicted velocity $\hat{\sigma}$ (right)}.
\label{fig:NavModelsPreds_mean}
\end{figure*}

In case of $\mathcal{M}_0$, the diversity from perception prediction samples improves the performance. Interestingly, the performance difference with other models is not significant. This situation can make us wonder if an uncertainty estimation is needed along the whole pipeline. Nonetheless, we believe that performance similarity is rooted in how we use our model predictions and uncertainties. The control output is computed by taking the mean and variance of the policy ensemble mixture, and only the mean values are passed to the UAV control. However, the multimodal predictions in Fig. \ref{fig:NavModelsPreds_mean} show that admitting perception uncertainty (samples) at the input of the control component permits the representation of ambiguity in the predictions. Hence, a proper use of predictions and associated uncertainties is needed. For example, in a bi-modal predictive distribution at the output, we can use the modes (i.e., distribution peaks) instead of the expected value to avoid sub-optimal control decisions (e.g., near distribution valleys).



\begin{figure}[t!]
\centering
    \subfloat[\centering Visible gate sub-dataset]
        {
        \includegraphics[width=0.97\linewidth]{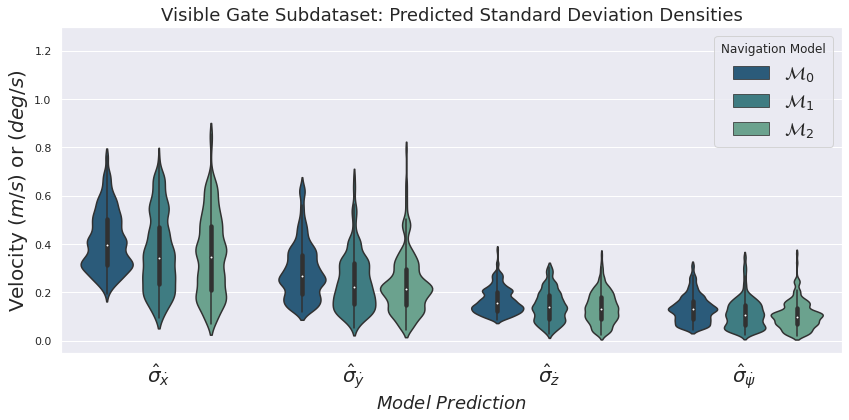}
        }
    
    \subfloat[\centering No visible gate sub-dataset]
        {
        \includegraphics[width=0.97\linewidth]{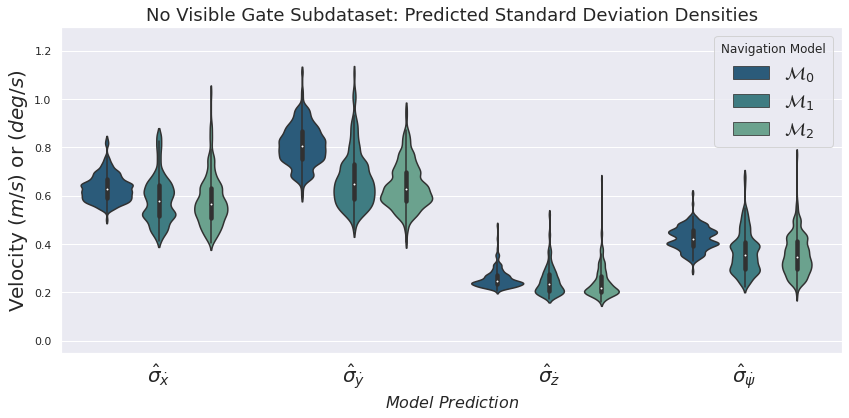}
        }
    
    \subfloat[\centering Multiple gates visible sub-dataset]
        {
        \includegraphics[width=0.97\linewidth]{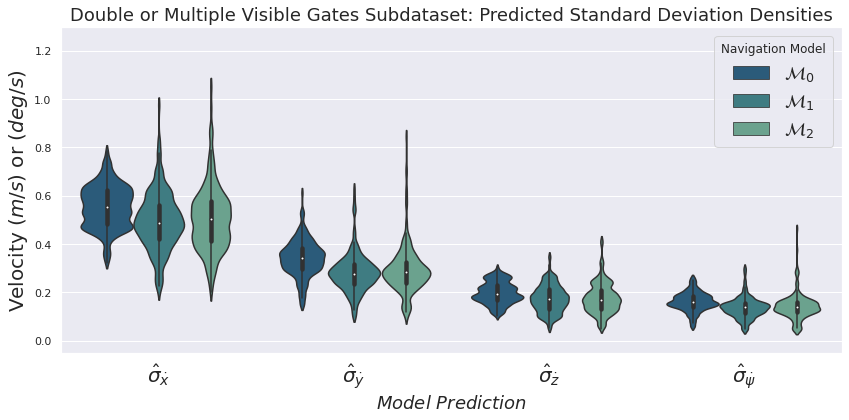}
        }
\caption{Navigation model standard deviation ($\hat{\sigma}$) prediction comparison}
\label{fig:NavModelsPreds_std}
\end{figure}

In the context of RQ3, Fig. \ref{fig:NavModelsPreds_std} shows the estimated uncertainty densities ($\hat{\sigma}$) for each velocity command at the output of the system, using the images from the generated datasets. In this case, $\mathcal{M}_0$ allows higher uncertainty estimates while reducing the dispersion in the sub-datasets from each situation. Fig. \ref{fig:NavModelsPreds_mean} shows $\mathcal{M}_0$ predictions at the output of the perception ($\mathbf{z}$) and control ($\hat{\mu}$, $\hat{\sigma}$) components. Predictions are made using the three sample images from Fig. \ref{subfig:scenes}, using the LVS and CPS to estimate the densities.

$\mathcal{M}_0$ perception and control outputs show high uncertainty (dispersion) values when a gate is not visible (mid-right). The $\hat{\mu}_{\dot{y}}$ density suggests that the UAV control predictions will follow the training dataset ($\mathcal{D}_c$) bias, rotating clockwise and moving to the right when no gate is in-front. Interestingly, the predicted densities in the bottom plots show that $\mathcal{M}_0$ is able to represent the ambiguity in the input, i.e. sample image with two gates. The predicted densities have a multimodal distribution (two peaks) for $\hat{\mu}_{\dot{y}}$ and $\hat{\sigma}_{\dot{y}}$ commands. Further, the predicted densities for the latent vector $\mathbf{z}$ show that the uncertainty from perception outputs is different for each type of sample, which is suitable for the early detection of anomalies based on uncertainty information. In addition, detecting multi-modality in prediction distributions can help expressing situations where decisions must be made.



\subsection{Dynamic Dependability Management using Uncertainty from DNN-Based Systems}

Based on the results and observations in the previous sub-sections, uncertainty propagation through a DNN-based can impact downstream component predictions and their performance. Thus, using uncertainty information to improve system dependability or safety can be a challenging task. For example, building monitoring functions based on uncertainty information is no simple task. The uncertainty intervals we observed for different situations present overlaps that can lead to false-positive or false-negative verdicts. Moreover, the multi-modal nature of some predictions under specific conditions or scenes demands knowledge of multiple intervals around the monitored uncertainty value. Therefore dependable and safe automated systems require more than a simple composition of predicates around some confidence measures.

\begin{figure}[!t]
	\centering
	\includegraphics[width=0.98\linewidth]{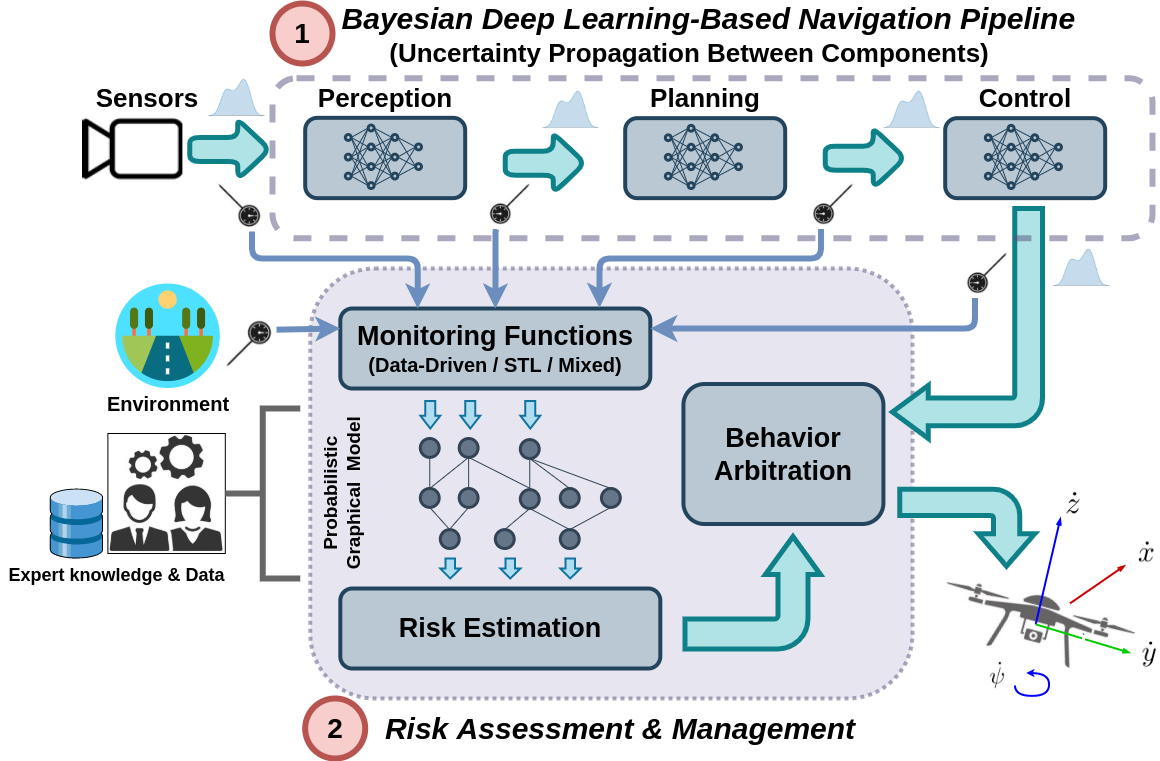}
	\caption{Runtime risk assessment \& management framework}
	\label{fig:runtime-risk-fw}
\end{figure}

Towards building dependable autonomous systems, we propose to align with previous frameworks that leverage perception uncertainty (cf. subsection \ref{subsec:uncertainty-dependability-fw}). However existing frameworks for system dependability do not consider the impact of uncertainty propagation in uncertainty-aware systems. To overcome these new challenges, we propose to capture and use uncertainty beyond perception and consider as well the uncertainty from downstream components along the navigation pipeline, as presented in~Fig.\ref{fig:runtime-risk-fw} \circled{1}. Our approach for dynamic dependability management takes inspiration from \cite{moosbrugger2017r2u2} and focuses on safety. Therefore, we propose an architecture for dynamic risk assessment and management where we devise three functional blocks, as shown in Fig. \ref{fig:runtime-risk-fw} \circled{2}: Monitoring functions, risk estimation and behavior arbitration modules.



\subsubsection{Monitoring Functions}

Monitoring is a widely-known dependability technique for runtime verification intended to track system variables (e.g. component inputs and outputs). In the automotive domain, SOTIF and ISO26262 suggest the use of monitoring functions as a solution for error detection in hardware and software components \cite{mohseni2019practical}. Monitoring functions are designed using a set of rules, based on a model of the system and its environment, and the properties they should guarantee. Hence, monitors basically perform a binary classification task to check if a property holds or not. 

Designing monitoring functions for ML components is different given the probabilistic nature of the outputs and the difficulty in specifying the component behavior at design time. For ML-based components in general, typical monitoring function tasks include Out-of-Distribution (OoD) detection or Out-of-Boundary (OoB) detection and can be implemented with rules, data-driven methods or a mix of both.


\subsubsection{Probabilistic Inference for Risk Assessment}

To enable dynamic uncertainty-aware reasoning and provide context to risk estimates, we propose to use Bayesian networks. Following the methodology described in \cite{kabir2019applications}, BNs for risk assessment and safety can be constructed using a combination of expert domain knowledge and data. The experts provide a model of causal relations and can have support from traditional dependability analysis (e.g., fault tree analysis) to build the BN structure while system data is used to provide the conditional probabilities between random variables.

In our framework, the BN of the system can receive the predictions from components in the pipeline (probability distributions) and the verdicts from monitoring functions applied to system sensors, component predictions, and relevant environmental variables. The output of the BN is represented by all the critical events identified by experts. Hence, during inference, the BN estimates the probability of a critical event, which is used along with its severity to compute the system's risk at runtime \cite{eggert2018risk}. Though we focus on risk assessment, in a general way the output of BNs can be any assurance measure variables linked to dependability attributes \cite{asaadi2020quantifying}. Further, the BN should handle \textit{uncertain evidence} \cite{mrad2015explication} to preserve the probabilistic nature of component and monitor predictions.

\subsubsection{Behavior Arbitration}
The last building block in our framework aims at keeping the system in a safe state by taking or discarding navigation pipeline predictions. Safe decisions must be made in the presence of high-risk values in a given context caused by erroneous component predictions or associated uncertainties and external environmental variables. To this end, we propose using Behavior Trees (BTs) to adopt different system behaviors while facing high-risk situations. BTs are sophisticated modular decision-making engines for reactive and fault-tolerant task execution \cite{colledanchise2021implementation}. Compositions of BTs can preserve safety and robustness properties \cite{colledanchise2014behavior} and are widely adopted tools in robotics. In the context of our system, we can have a dedicated behavior to search for a gate when we detect that there are no gates in the UAV FoV. This behavior will put the system back into a state where the levels of uncertainty do not represent a risk.

\section{Conclusion}
We presented a method to capture and propagate uncertainty along a navigation pipeline implemented with Bayesian deep learning components for UAV aerial navigation. We analyzed the effect of uncertainty propagation regarding system component predictions and performance. Our experiments show that our approach to capturing and propagating uncertainty along the system can provide valuable predictions for decision-making and identifying situations that are critical for the system. However, proper use and management of component predictions and uncertainty estimates are needed to create dependable and highly-performant systems. In this sense and based on our observations, we also proposed a framework for system dependability management using system uncertainty and focused on safety. In future work, we aim to implement our proposed dependability framework and explore sampling-free methods \cite{charpentier2021natural} for uncertainty estimation to reduce the computational budget and memory footprint in our approach.




\ifCLASSOPTIONcompsoc
  \section*{Acknowledgments}
\else
  \section*{Acknowledgment}
\fi
This work has received funding from the COMP4DRONES project, under ECSEL Joint Undertaking (JU) grant agreement N\textdegree  826610. The ECSEL JU receives support from the European Union's Horizon 2020 research and innovation programme and from Spain, Austria, Belgium, Czech Republic, France, Italy, Latvia, Netherlands.

\bibliographystyle{IEEEtran}
\bibliography{my_bib}


\end{document}